\documentclass[11pt, a4paper, copyright, gdm]{google}

\usepackage[authoryear, sort&compress, round]{natbib}
\uselogo{} 

\usepackage{booktabs}
\usepackage{tcolorbox}
\usepackage{subcaption}
\usepackage{graphicx} 
\usepackage{algorithm}
\usepackage{algorithmic}

\def\eqref#1{equation~\ref{#1}}

\def\1{\bm{1}}

\DeclareMathAlphabet{\mathsfit}{\encodingdefault}{\sfdefault}{m}{sl}
\SetMathAlphabet{\mathsfit}{bold}{\encodingdefault}{\sfdefault}{bx}{n}

\def\gD{{\mathcal{D}}}

\def\gL{{\mathcal{L}}}

\newcommand{\E}{\mathbb{E}}

\definecolor{chosenblue}{rgb}{0.208,0.463,0.725}

\definecolor{rejectedred}{rgb}{0.682,0.094,0.094}

\title{MUSIC: MUlti-Step Instruction Contrast for Multi-Turn Reward Models}

\reportnumber{} 

\renewcommand{\today}{}

\author[1,*]{Wenzhe Li}
\author[2]{Shujian Zhang}
\author[2]{Wenxuan Zhou}
\author[2]{John Lambert}
\author[1]{Chi Jin}
\author[2]{Andrew Hard}
\author[2]{Rajiv Mathews}
\author[2]{Lun Wang}

\affil[1]{Princeton University}
\affil[2]{Google DeepMind}
\affil[*]{Work done as a student researcher at Google DeepMind.}

\begin{abstract}
  Evaluating the quality of multi-turn conversations is crucial for developing capable Large Language Models (LLMs), yet remains a significant challenge, often requiring costly human evaluation. Multi-turn reward models (RMs) offer a scalable alternative and can provide valuable signals for guiding LLM training. While recent work has advanced multi-turn \textit{training} techniques, effective automated \textit{evaluation} specifically for multi-turn interactions lags behind. We observe that standard preference datasets, typically contrasting responses based only on the final conversational turn, provide insufficient signal to capture the nuances of multi-turn interactions. Instead, we find that incorporating contrasts spanning \textit{multiple} turns is critical for building robust multi-turn RMs. Motivated by this finding, we propose \textbf{MU}lti-\textbf{S}tep \textbf{I}nstruction \textbf{C}ontrast (MUSIC), an unsupervised data augmentation strategy that synthesizes contrastive conversation pairs exhibiting differences across multiple turns. Leveraging MUSIC on the Skywork preference dataset, we train a multi-turn RM based on the Gemma-2-9B-Instruct model. Empirical results demonstrate that our MUSIC-augmented RM outperforms baseline methods
  , achieving higher alignment with judgments from advanced proprietary LLM judges on multi-turn conversations, crucially, without compromising performance on standard single-turn RM benchmarks.
\end{abstract}

\begin{document}

\maketitle

\section{Introduction}

The ability of Large Language Models (LLMs) to engage in coherent, multi-turn conversations is a hallmark of advanced AI systems~\citep{turing1950computing}. While recent LLMs demonstrate remarkable proficiency in single-turn instruction following and short dialogues~\citep{ouyang2022training,adler2024gpt,team2023gemini}, extending this capability to complex, long-horizon interactions remains a critical frontier~\citep{zheng2023judging,abdulhai2023lmrl,he2024multi, deshpande2025multichallenge}. Significant effort has focused on developing Reinforcement Learning from Human Feedback (RLHF) techniques tailored for multi-turn dynamics~\citep{zhou2024archer,gao2024regressing,shi2024direct,shani2024multi,he2025rubric,abdulhai2025consistently,jiang2025aligning}, aiming to improve conversational performance beyond standard single-turn RLHF methods.

Despite advances in multi-turn \textit{training}, robust automated \textit{evaluation} of these interactions presents a persistent challenge. High-quality, model-based evaluators, or specifically reward models (RMs), are crucial, serving not only as direct performance metrics but also providing signals during training and inference~\citep{lambert2024rewardbench,malik2025rewardbench}. However, evaluating multi-turn conversations is fundamentally more complex than single-turn assessment. It requires judging not only the response quality at each turn but also inter-turn properties like coherence, consistency, and effective use of conversational history~\citep{deshpande2025multichallenge,he2024multi}. Consequently, training powerful multi-turn RMs typically necessitates large volumes of high-quality preference data reflecting these nuances~\citep{wang2024self,wang2024helpsteer2,liu2024skywork}.

Acquiring such data via human annotation is prohibitively expensive. Comparing two lengthy conversations, potentially differing subtly across multiple turns, is significantly more demanding and time-consuming than annotating single-turn preferences~\citep{deshpande2025multichallenge}. As a result, widely used preference datasets~\citep{bai2022training,ganguli2022red,pmlr-v162-ethayarajh22a,cui2023ultrafeedback,liu2024skywork} often contain predominantly single-turn pairs or multi-turn pairs where the difference is confined to the final turn. While practical for efficient data collection, this data characteristic may limit the ability of RMs trained on them to capture holistic conversational quality. 
This motivates our central research question:
\begin{center}
    \textit{Can we develop a scalable approach to synthesize contrastive data spanning multiple turns to train more effective multi-turn RMs?}
\end{center}

To address this data gap, we propose \textbf{MU}lti-\textbf{S}tep \textbf{I}nstruction \textbf{C}ontrast (MUSIC), an unsupervised data augmentation strategy designed to generate contrastive conversation pairs with meaningful quality differences distributed across multiple turns, without human annotation. By introducing controlled variations of instructions during the generation process, one conversation in the pair is constructed to be qualitatively better (e.g., more consistent, exhibiting better instruction following) than the other \textit{across multiple turns}. This creates contrastive data specifically highlighting multi-turn phenomena where the quality distinction is woven throughout the conversation. MUSIC can be readily applied to augment existing preference datasets, enriching them with multi-turn contrast signals.

We demonstrate the efficacy of MUSIC by applying it to the Skywork preference dataset~\citep{liu2024skywork} and subsequently fine-tuning a Gemma-2-9B-Instruct model on this augmented data. Our experiments show that the resulting MUSIC-augmented RM maintains strong performance on standard single-turn benchmarks like RewardBench~\citep{lambert2024rewardbench}. More importantly, compared to baseline models trained without MUSIC, our RM exhibits higher agreement with judgments from the advanced Gemini 1.5 Pro model when assessing the quality of multi-turn conversations.

Our contributions are threefold:
\begin{enumerate}
    \item We identify a critical limitation in standard preference datasets for training multi-turn RMs: the predominant focus on final-turn contrasts, which hinders the learning of holistic conversational quality assessment.
    \item We propose MUSIC, a scalable, unsupervised method to synthesize contrastive conversation pairs with meaningful quality differences spanning multiple turns, directly addressing the identified data gap.
    \item We demonstrate empirically that RMs trained with MUSIC achieve improved alignment with advanced LLM judges on multi-turn tasks, without sacrificing performance on single-turn benchmarks, validating the effectiveness of our approach.
\end{enumerate}

\begin{figure*}[t]
  \centering
  \includegraphics[width=\textwidth]{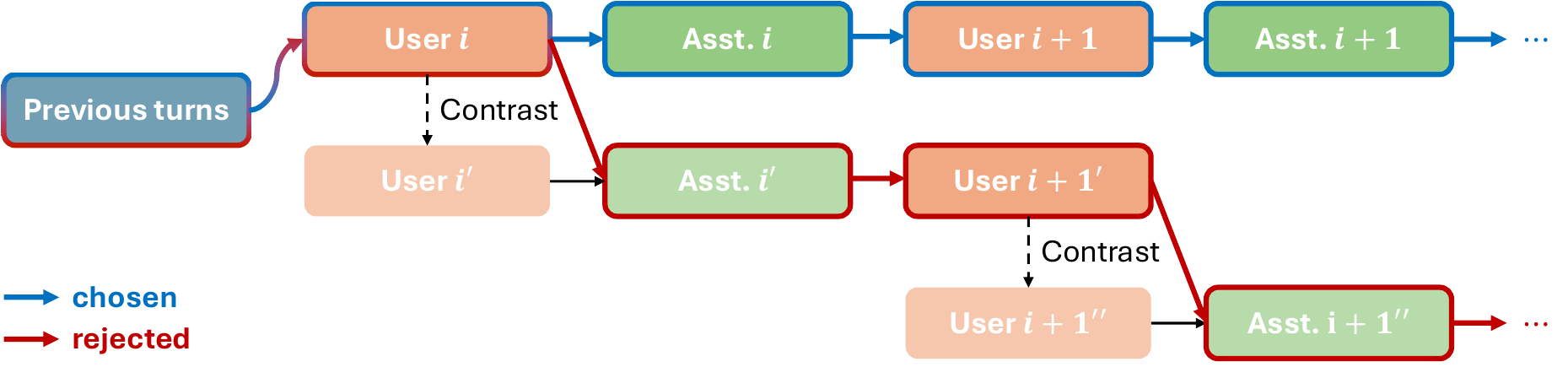}
  \caption{Overview of the MUSIC data augmentation procedure. Given seed contexts from existing datasets, we generate multi-turn rollouts where LLM simulators generate contrastive pairs, and use a contrastive instruction prompt to induce quality degradation in the rejected branch. The augmented preference pairs are used to train a multi-turn reward model along with the original dataset. Black arrows represent ephemeral changes that are provided to the assistant once, but not persisted. 
  For each augmented pair, the \textcolor{chosenblue}{chosen} example consists of turns with \textcolor{chosenblue}{blue} borders, while the \textcolor{rejectedred}{rejected} example consists of turns with \textcolor{rejectedred}{red} borders.
  }
  \label{fig:music}
\end{figure*}

\section{Related Work}

\paragraph{Preference Learning and Reward Modeling.}
Aligning LLMs with human values has evolved significantly since the foundational frameworks of Reinforcement Learning from Human Feedback (RLHF) were established~\citep{christiano2017deep,ziegler2019fine}. The standard pipeline relies on learning a reward model (RM) from human preferences to guide policy optimization~\citep{ouyang2022training, bai2022training}. While alternatives like Direct Preference Optimization (DPO)~\citep{rafailov2023direct} bypass explicit reward modeling, RMs remain essential for scalable oversight, rejection sampling, and guiding search~\citep{lambert2024rewardbench}, especially in domains without verifiable rewards. Recent literature on RMs has bifurcated into two distinct streams:
\begin{itemize}
    \item \textbf{Outcome Reward Models (ORMs):} These models typically assign a single scalar score to an entire LLM generation (e.g., a full response or conversational turn)  based on its overall quality~\citep{liu2024skywork,wang2024helpsteer2,cobbe2021training,wang2024arithmetic}. They are widely used for general instruction following, dialogue, and reasoning tasks.
    \item \textbf{Process Reward Models (PRMs):} These provide denser supervision by evaluating intermediate steps within a generation process, such as individual reasoning steps in mathematical proofs or lines of code~\citep{uesato2022solving,lightman2023let,wang2023math}, and recently extending to more general domains~\citep{zeng2025versaprm,yin2025dynamic}. However, PRMs require more fine-grained annotations and thus are more expensive to train compared to ORMs.
\end{itemize}
Our work focuses on enhancing ORMs for open-ended \textit{multi-turn conversations}, where "steps" are conversational turns and the quality signal is often implicit and distributed rather than discrete and verifiable. While standard ORM training often relies on preference data where contrasts are localized (e.g., single-turn differences or final-turn edits in dialogues), MUSIC acts as a data augmentation technique. It synthesizes preference pairs where the quality difference is intentionally distributed across multiple turns. By training standard ORM architectures on MUSIC-augmented data, we aim to improve their ability to capture holistic multi-turn properties like coherence and consistency, which are often underspecified by conventional preference datasets and distinct from the step-level focus of PRMs.

\paragraph{Multi-Turn Alignment.}
Extending alignment to multi-turn interactions introduces significant complexity due to long-term credit assignment challenges~\citep{abdulhai2023lmrl}. Early dialogue systems often use handcrafted reward functions based on heuristics~\citep{li2016deep} for RL on small-scale models, while more recent approaches investigate RL techniques on LLM tailored for multi-turn alignment, including but not limited to hierarchical RL~\citep{zhou2024archer}, value-based~\citet{jiang2025aligning} and self-play or multi-agent~\citet{shani2024multi,wu2025aligning} methods. However, these advanced policy optimization methods depend critically on robust reward signals. While existing multi-turn benchmarks~\citep{he2024multi,deshpande2025multichallenge,he2025rubric} leverage human annotations or rubric-based methods for evaluation, such efforts are often costly and not scalable for training.
Our work complements this line of work by enhancing the underlying reward models through MUSIC, thereby improving the overall multi-turn alignment process.

\paragraph{Synthetic Data for Alignment.}
The scarcity of high-quality human annotations has driven a shift toward synthetic data generation. Recent work demonstrates that LLMs could generate their own fine-tuning data~\citep{wang2023self, dubois2023alpacafarm} and provide feedback signals for improvement~\citep{chen2024self,yuan2024self}. This paradigm is also used to generate multi-turn conversations for LLM training more recently~\citep{wu2025instruct,yin2025aligning}.
Unlike these methods, which primarily focus on generating data for SFT or RL, MUSIC focuses specifically on synthesizing \textit{contrastive preference pairs} to train a multi-turn RM. We automate the creation of chosen and rejected trajectories by injecting controlled noise, thereby providing the necessary discriminative signals.

\section{Method}

We introduce \textbf{MU}lti-\textbf{S}tep \textbf{I}nstruction \textbf{C}ontrast (MUSIC), a scalable, unsupervised method for synthesizing contrastive conversation pairs that exhibit meaningful quality differences across multiple turns. This synthesized data is designed to augment existing preference datasets, enabling the training of more effective multi-turn RMs.
The core process involves three stages:
\begin{itemize}
    \item \textbf{Initialization:} We sample conversational prefixes (seed contexts) from an existing multi-turn dataset to initiate the augmentation process.
    \item \textbf{Multi-turn Rollouts with MUSIC:} Starting from each seed context, we employ LLM-based user and assistant simulators to generate paired conversations. Crucially, at each turn, a contrastive instruction prompt guides the assistant simulator to produce a lower-quality response for one conversation in the pair.
    \item \textbf{Multi-turn RM Training:} The conversation pairs generated by MUSIC are combined with the original preference data. A multi-turn RM is then trained on this augmented dataset using standard preference learning techniques.
\end{itemize}

In this section, we first review the preliminaries for training model-based RMs, and then describe each stage of our pipeline in detail. 

\begin{algorithm}[t]
\caption{\textbf{MU}lti-\textbf{S}tep \textbf{I}nstruction \textbf{C}ontrast (MUSIC) Data Generation}
\label{alg:music}
\begin{algorithmic} 
\REQUIRE Seed conversation context $C_{\text{prefix}}$, LLM user simulator $M_u$, LLM assistant simulator $M_a$, max simulation turns $T$, instruction contrast prompt $\texttt{Contrast}(\cdot)$
\STATE Initialize $C_{\text{chosen}} \leftarrow C_{\text{prefix}}$, $C_{\text{rejected}} \leftarrow C_{\text{prefix}}$
\FOR{$t = 1$ \TO $T$}
    \STATE Generate next user utterance: $u_t^{\text{chosen}} \leftarrow M_u(C_{\text{chosen}})$, $u_t^{\text{rejected}} \leftarrow M_u(C_{\text{rejected}})$ 
    \STATE Generate chosen assistant response: $a_t^{\text{chosen}} \leftarrow M_a(C_{\text{chosen}} \oplus u_t^{\text{chosen}})$ 
    \STATE Generate rejected assistant response: $a_t^{\text{rejected}} \leftarrow M_a(C_{\text{rejected}} \oplus \texttt{Contrast}(u_t^{\text{rejected}}))$ 
    \STATE Append turn to the context:
    \begin{align*}
        &C_{\text{chosen}} \leftarrow C_{\text{chosen}} \oplus (u_t^{\text{chosen}}, a_t^{\text{chosen}}), \\ &C_{\text{rejected}} \leftarrow C_{\text{rejected}} \oplus (u_t^{\text{rejected}}, a_t^{\text{rejected}})
    \end{align*}
\ENDFOR
\RETURN $(C_{\text{chosen}}, C_{\text{rejected}})$
\end{algorithmic}
\end{algorithm}

\subsection{Preliminaries}
We focus on ORMs, where the model $R_\theta$ maps a conversation (or parts thereof) to a scalar score~\citep{liu2024skywork,wang2024helpsteer2}. Training typically involves maximizing the log-likelihood of observing human preferences under the Bradley-Terry (BT) model~\citep{bradley1952rank}:
\begin{align}
\label{eq:bt_loss}
    &\gL(\theta,\gD) = 
    \E_{C_{\text{chosen}},C_{\text{rejected}}\sim\gD}
    \log\sigma\left (R_\theta(C_{\text{chosen}}) - R_\theta(C_{\text{rejected}})\right )
\end{align}
where $\gD$ is a preference dataset of conversation pairs $(C_{\text{chosen}},C_{\text{rejected}})$, and $\sigma$ is the sigmoid function, respectively. In practice, $R_\theta$ is often implemented by fine-tuning a pre-trained or instruction-tuned LLM, adding a linear layer to map a representation (e.g., the last-layer hidden state of the final token) to the scalar reward score. 

\subsection{Initialization}
\label{sec:init}
We assume access to an existing multi-turn preference dataset $\gD=\{(C^{(i)}_{\text{chosen}},C^{(i)}_{\text{rejected}})\}_{i=1}^N$. As noted earlier, such datasets~\citep{bai2022training,ganguli2022red,liu2024skywork} often contain pairs differing only in the final turn, providing limited signal for multi-turn phenomena. However, the initial turns often represent valid, human-generated conversational trajectories. We leverage this by sampling seed contexts from $\gD$. Specifically, for conversation $C^{(i)}$ in the dataset of $H$ turns, we sample a turn index $h \sim U(1, H)$ uniformly at random and extract the first $h$ turns as the seed context $C_{\text{prefix}} = C^{(i)}_{1:h}$. This approach balances the reuse of high-quality human-curated conversational prefixes with the generation of novel multi-turn contrasts via MUSIC.

\subsection{Multi-turn Rollouts with MUSIC}

Given a set of seed contexts, we apply the MUSIC algorithm (Algorithm~\ref{alg:music}) to generate contrastive conversation pairs $\gD_{\text{MUSIC}}$. This process simulates multi-turn interactions using LLMs as proxies for both the user ($M_u$) and the assistant ($M_a$), inspired by work on generative agents~\citep{park2023generative,park2024generative}.

The core idea of MUSIC is to introduce controlled quality degradation in one branch of the simulated conversation pair at each turn. This is achieved via the instruction contrast prompt, $\texttt{Contrast}(\cdot)$. For the \textit{chosen} conversation path $C_{\text{chosen}}$, the simulated assistant $M_a$ responds directly to the simulated user's utterance $u_t^{\text{chosen}}$. For the \textit{rejected} path $C_{\text{rejected}}$, however, the user's utterance $u_t^{\text{rejected}}$ is first transformed by $\text{Contrast}(\cdot)$ into a modified instruction, which prompts $M_a$ to generate a response $a_t^{\text{rejected}}$ that is intentionally suboptimal relative to the original user utterance $u_t^{\text{rejected}}$ (e.g., less helpful, inconsistent with previous turns, or failing to follow a specific constraint). As shown in Figure~\ref{fig:music}, the instruction contrast prompt implicitly guides the assistant to generate responses through ephemeral modifications, ensuring the rejected trajectory remains coherent yet qualitatively inferior to its chosen counterpart. The design details of $\texttt{Contrast}(\cdot)$ are provided in Appendix~\ref{sec:contrast_prompt}, drawing inspiration from~\citep{wang2024self}.

By repeating this process for $T$ turns, MUSIC generates paired conversations $(C_{\text{chosen}}, C_{\text{rejected}})$ where $C_{\text{chosen}}$ is superior by construction, and the quality difference is distributed across multiple turns rather than being localized. This yields preference data specifically designed to train RMs sensitive to multi-turn conversational dynamics.


\subsection{Multi-turn RM Training}

After generating the MUSIC dataset $\gD_{\text{MUSIC}}$, we create the final augmented training dataset $\gD_{\text{aug}}=\gD\cup\gD_{\text{MUSIC}}$. We then train our multi-turn RM $R_\theta$ on $\mathcal{D}_{\text{aug}}$ by optimizing the BT loss objective in Equation~\ref{eq:bt_loss}. We train for a small number of epochs (e.g., less than two) to mitigate potential overfitting to the combined dataset. The resulting RM $R_\theta$ is expected to have improved sensitivity to multi-turn conversational properties due to its exposure to the contrastive examples synthesized by MUSIC.

\section{Experiments}
Our experiments are designed to investigate the efficacy of MUSIC by addressing the following research questions: 
\textbf{(a)} Does MUSIC improve the effectiveness of RMs for assessing multi-turn conversations?
\textbf{(b)} Does augmenting training data with MUSIC negatively impact the RM's performance on standard single-turn RM benchmarks?

To answer \textbf{(a)}, we evaluate the performance of a MUSIC-augmented RM against a baseline RM (trained without MUSIC) in a multi-turn Best-of-N (BoN) inference task. This task requires the RM to iteratively select the best response from $N$ candidates generated by an assistant LLM at each turn of a conversation. The quality of the resulting multi-turn conversations serves as a proxy for the RM's effectiveness.
To answer \textbf{(b)}, we evaluate both RMs on RewardBench~\citep{lambert2024rewardbench}, a standard benchmark primarily focused on single-turn evaluation capabilities.

\phantomsection
\subsection{Experimental Setup}
\paragraph{Dataset Construction.}
We use \href{https://huggingface.co/datasets/Skywork/Skywork-Reward-Preference-80K-v0.2}{Skywork-Reward-Preference-80K-v0.2} as the RM training dataset as it is used to train several state-of-the-art RMs~\citep{dorka2024quantile,skyworkcritic2024,liu2024skywork}. 
This dataset is representative of standard preference data, containing mostly single-turn pairs and multi-turn pairs differing only in the final turn, making it a suitable candidate for augmentation with MUSIC.
Specifically, we filter the dataset to include only dialogues with at most five turns and uniformly sample the seed contexts as described in Section~\ref{sec:init}. For MUSIC augmentation, we use Gemini 1.5 Pro as both user and assistant simulators with distinct prompts (see Appendix~\ref{sec:user_prompt} and~\ref{sec:assistant_prompt}), and set the maximum simulation turns $T=5$. Both $\gD$ and $\gD_{\text{MUSIC}}$ are preprocessed by filtering out conversations exceeding 2048 tokens (the maximum sequence length for training). Our final datasets consist of approximately 73k pairs from the original Skywork-Reward-Preference-80K-v0.2 dataset and 31k pairs from the MUSIC augmentation.

\paragraph{Training Details.} 

We fine-tune Gemma-2-9B-Instruct~\citep{team2024gemma} to create our RMs. A linear layer is added on top of the transformer's final hidden state output to produce a scalar reward score. We train two main models:
\begin{enumerate}
    \item \textbf{Baseline RM:} Trained on $\mathcal{D}$.
    \item \textbf{MUSIC-Augmented RM:} Trained on $\mathcal{D}_{\text{aug}}$.
\end{enumerate}
Both models are trained using the AdamW optimizer~\citep{loshchilov2017decoupled} with a learning rate of $2 \times 10^{-6}$, a global batch size of 64, and a maximum sequence length of 2048. We use a cosine learning rate decay schedule and train for 2500 steps to minimize the Bradley-Terry loss (Equation~\ref{eq:bt_loss}).

\paragraph{Evaluation Details.}
We compare the Baseline RM and the MUSIC-Augmented RM on the following tasks:
\begin{enumerate}
    \item \textbf{Multi-Turn Best-of-N (BoN) Inference:} Best-of-N is an effective approach to leverage single-turn RMs to improve LLMs' single-turn capability~\citep{brown2024large,wu2024inference,snell2024scaling}. 
    This task assesses the RM's ability to guide an LLM assistant towards generating higher-quality multi-turn conversations. We simulate interactions between a user (Gemini 1.5 Pro) and an assistant (Gemma-2-9B-Instruct). Conversations are initiated using 1000 prompts sampled from subsets of Anthropic HH~\citep{bai2022training} and UltraInteract~\citep{yuan2024advancing}, following~\citep{gao2024regressing}. At each of $H=3$ turns, the assistant generates $N \in \{2, 4, 8\}$ candidate responses at a fixed temperature. The RM being tested selects the response with the highest score, which is then used to continue the conversation. The number of turns $H=3$ was chosen to accommodate the 2048 context length of the RMs and assistant. After $H$ turns, the quality of the full conversation generated using the MUSIC-Augmented RM is compared against the conversation generated using the Baseline RM. We use Gemini 1.5 Pro as an LLM judge, prompting it to select the better conversation based on criteria adapted from~\citep{zheng2023judging} (prompt in Appendix~\ref{sec:evaluator_prompt}). To mitigate positional bias, each pair of conversations is evaluated twice with the order swapped, and we report the average winrate. We also compare against greedy decoding from the assistant as a reference.
    \item \textbf{RewardBench:} 
    To assess single-turn performance, we evaluate both RMs on RewardBench~\citep{lambert2024rewardbench}. Following the standard protocol, we report pairwise accuracy across its four main categories (Chat, Chat Hard, Safety, Reasoning) and the overall average accuracy.
\end{enumerate}

\subsection{Results on Multi-Turn Best-of-N Inference}




\begin{figure*}[t]
    \centering
    \includegraphics[width=\textwidth]{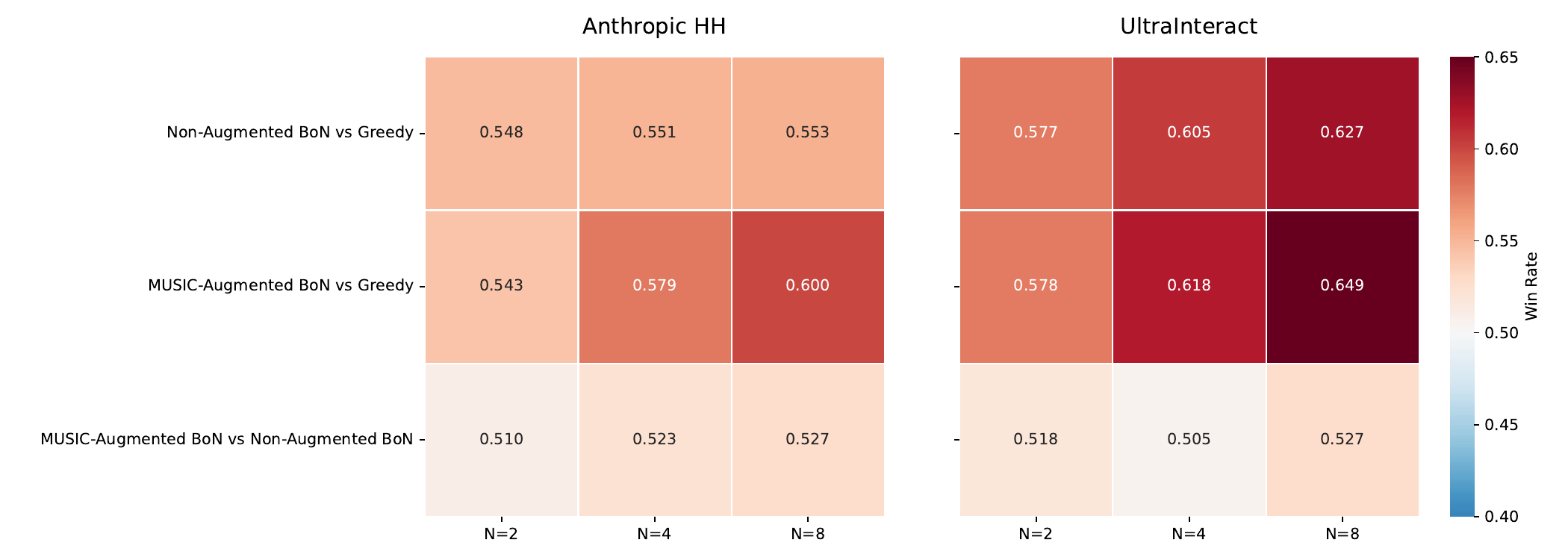} 
    \caption{
    Winrates comparing conversations generated via Best-of-N ($N \in \{2, 4, 8\}$) guided by the MUSIC-Augmented RM versus the Baseline (non-augmented) RM, evaluated by Gemini 1.5 Pro on subsets of Anthropic HH and UltraInteract. Comparisons against greedy decoding are also shown.
    } 
    \label{fig:winrate}
\end{figure*}

Figure~\ref{fig:winrate} presents the winrates from the multi-turn BoN evaluation. We compare conversations generated using BoN guided by the MUSIC-Augmented RM against those guided by the Baseline RM, as judged by Gemini 1.5 Pro. For reference, we also include comparisons against greedy decoding from the assistant LLM.

Across both the Anthropic HH and UltraInteract initial prompts, the results consistently demonstrate that conversations guided by the MUSIC-Augmented RM are preferred over those guided by the Baseline RM. Furthermore, the performance gap generally widens as $N$ increases, indicating that the MUSIC-Augmented RM effectively leverages the stronger candidate pool provided by larger $N$. Both BoN methods outperform the greedy baseline substantially. This provides strong evidence for research question (a): \textbf{MUSIC successfully enhances the RM's ability to identify and promote higher-quality multi-turn interactions}, leading to demonstrably better conversational outputs as judged by an advanced LLM.

\subsection{Results on RewardBench}

\begin{table*}[t] 
  \centering 
  \caption{
RewardBench accuracy (\%) results. We compare the RM trained on the original Skywork dataset and the RM trained on the MUSIC-augmented dataset. Both use Gemma-2-9B-Instruct as the base model.
  } 
  \label{tab:reward_bench} 
  \begin{tabular}{lccccc} 
    \toprule
    \textbf{Model} & \textbf{Overall} & \textbf{Chat} & \textbf{Chat Hard} & \textbf{Safety} & \textbf{Reasoning} \\
    \midrule
    Gemma-2-9B-Instruct w/ Skywork & 85.7 & \textbf{91.9} & 83.8 & 88.4 & 78.6 \\
    Gemma-2-9B-Instruct w/ Skywork + MUSIC & \textbf{87.2} & 91.6 & \textbf{85.1} & \textbf{89.7} & \textbf{82.5} \\
    \bottomrule
  \end{tabular}
\end{table*}

Table~\ref{tab:reward_bench} shows the performance of the Baseline and MUSIC-Augmented RMs on RewardBench. Addressing research question (b), we observe that \textbf{augmenting the training data with MUSIC does not sacrifice single-turn evaluation performance}. In fact, the MUSIC-Augmented RM achieves slightly better or comparable accuracy across the Chat, Chat Hard, and Safety categories.

Surprisingly, we observe a notable improvement (+3.9\%) in the Reasoning category for the MUSIC-Augmented RM. While MUSIC synthesizes multi-turn conversational data and is not explicitly designed for single-turn reasoning tasks, this suggests a potential positive transfer. We hypothesize that exposure to coherent, logically structured multi-turn dialogues during training may implicitly enhance the RM's ability to assess reasoning steps, even when presented in single turns. Overall, these results indicate that \textbf{MUSIC not only improves multi-turn evaluation capabilities but does so without compromising, and potentially even slightly enhancing, performance on standard single-turn benchmarks}.

\section{Conclusion}
In this work, we addressed the challenge of evaluating multi-turn conversations by introducing \textbf{MU}lti-\textbf{S}tep \textbf{I}nstruction \textbf{C}ontrast (MUSIC), a scalable, unsupervised data augmentation technique. MUSIC synthesizes contrastive conversation pairs where quality differences are intentionally distributed across multiple turns, enriching standard preference datasets that often focus on final-turn contrasts. We demonstrated that training a multi-turn RM on a MUSIC-augmented dataset leads to improved performance in guiding multi-turn interactions, as measured by alignment with judgments from an advanced LLM judge in a Best-of-N setting. Crucially, these gains in multi-turn evaluation capability were achieved without compromising, and potentially even slightly enhancing, performance on standard single-turn benchmarks like RewardBench. Our results validate MUSIC as an effective strategy for training more robust multi-turn RMs, mitigating the need for expensive human annotation of complex conversational preferences.


\section{Limitations and Future Work}
\label{sec:limitations}

While promising, our work has several limitations that suggest avenues for future research.

\noindent\textbf{Reliance on LLM Simulators and Judges:} Both the MUSIC data generation process (using $M_u$ and $M_a$) and the primary multi-turn evaluation (BoN judged by Gemini 1.5 Pro) rely heavily on LLMs. While practical and scalable, these models may introduce their own biases or fail to capture the full spectrum of human conversational nuances and preferences. Future work could explore incorporating real human interactions or judgments, potentially through targeted human-in-the-loop refinement or evaluation on human-annotated multi-turn benchmarks, to further validate and potentially improve the approach.

\noindent\textbf{Conversation Length and Model Scale:} Our experiments were constrained by computational resources and model context windows, limiting MUSIC rollouts to $T=5$ turns and BoN evaluation to $H=3$ turns. The effectiveness of MUSIC for significantly longer conversations remains to be explored. Scaling the approach to larger base models with longer context windows is a natural next step, potentially unlocking benefits for evaluating more complex, extended dialogues.

Addressing these limitations represents promising directions for advancing automated evaluation of complex, multi-turn LLM interactions.

\newpage

\bibliography{main}

\newpage

\appendix

\section{Prompt Design}

\phantomsection
\subsection{User Simulator Prompt}
\label{sec:user_prompt}
The prompt template for the user simulator is adapted from~\citep{gao2024regressing,dubois2024length,rafailov2023direct}:

\begin{tcolorbox}[width=\linewidth, colback=white!95!black]
Below is a dialogue between the user and the assistant. Pretend you are the user in this conversation. What question would you ask next? 

\vspace{10pt}

\{\{previous turns\}\}

\vspace{10pt}

\#\#\# Instructions: 

FIRST, provide a justification of the question you want to ask. 

SECOND, on a new line, state only the question. 

Your response should use the format: 

Justification: 

Question: 
\end{tcolorbox}

\vspace{-15pt}

\phantomsection
\subsection{Assistant Simulator Prompt}
\label{sec:assistant_prompt}
For the assistant LLM, we directly follow the prompt template provided in~\citep{team2024gemma}:

\begin{tcolorbox}[width=\linewidth, colback=white!95!black]

<start\_of\_turn>user \\
\{\{1st turn instruction\}\}<end\_of\_turn> \\
<start\_of\_turn>model \\
\{\{1st turn response\}\}<end\_of\_turn> \\
<start\_of\_turn>user \\
\{\{2nd turn instruction\}\}<end\_of\_turn> \\
<start\_of\_turn>model \\
\{\{2nd turn response\}\}<end\_of\_turn> \\
\dots \\
<start\_of\_turn>user \\
\{\{last turn instruction\}\}<end\_of\_turn> \\
<start\_of\_turn>model
\end{tcolorbox}

\vspace{-15pt}

\phantomsection
\subsection{Instruction Contrast Prompt}
\label{sec:contrast_prompt}
The instruction contrast prompt is the core to synthesize turn-level differences in MUSIC. Inspired by~\citep{wang2024self}, we directly encode the instruction contrast prompt into the prompt for the assistant LLM to generate the rejected conversations in the preference pairs:

\begin{tcolorbox}[width=\linewidth, colback=white!95!black]
Below is a dialogue between the user and the assistant. Pretend you are the assistant in this conversation. 

\vspace{10pt}

\{\{previous turns\}\}

\vspace{10pt}

\#\#\# Instructions: 

FIRST, generate a modified instruction that is highly relevant but not semantically identical to the instruction above from the user in the last turn. 

SECOND, on a new line, generate a high-quality answer which is a good response to the modified instruction but not a good response to the original user question.

Your response should use the format: 

Modified Instruction: 

Answer: 
\end{tcolorbox}

\vspace{-15pt}

\phantomsection
\subsection{Evaluator Prompt}
\label{sec:evaluator_prompt}
We adapt the single-turn evaluation prompt from~\citep{zheng2023judging} to evaluate multi-turn conversations:

\begin{tcolorbox}[width=\linewidth, colback=white!95!black]
Please act as an impartial judge and evaluate the quality of the conversation between the user and two AI assistants displayed below. You should choose the assistant that follows the user's instructions and answers the user's questions better. Your evaluation should consider factors such as the helpfulness, relevance, accuracy, depth, creativity, and level of detail of their responses. Begin your evaluation by comparing the two conversations and provide a short explanation. Avoid any position biases and ensure that the order in which the conversations were presented does not influence your decision. Do not allow the length of the responses to influence your evaluation. Do not favor certain names of the assistants. Be as objective as possible. After providing your evaluation, output your final verdict by strictly following this format: "[[A]]" if assistant A is better, "[[B]]" if assistant B is better.
\end{tcolorbox}


\begin{tcolorbox}[width=\linewidth, colback=white!95!black]
[The Start of Assistant A's Conversation]

\{\{conversation A\}\}

[The End of Assistant A's Conversation]

\vspace{10pt}

[The Start of Assistant B's Conversation]

\{\{conversation B\}\}

[The End of Assistant B's Conversation]
\end{tcolorbox}

\end{document}